\documentclass{style/nseJournal}

\usepackage{subcaption}
\usepackage{booktabs}
\usepackage{multirow}
\usepackage{xcolor}
\usepackage{soul}
\usepackage{placeins}
\usepackage[numbers,sort&compress]{natbib}

\begin{document}

\title{Learning to Resolve Neutron Resonances with Fully Convolutional Neural Networks}

\addAuthor{\correspondingAuthor{Nataly R. Panczyk}}{a,b}
\correspondingEmail{npanczyk@umich.edu; radaideh@umich.edu}
\addAuthor{Athanasios Stamatopoulos}{b}
\addAuthor{Josef Svoboda}{b}
\addAuthor{\correspondingAuthor{Majdi I. Radaideh}}{a,c}

\addAffiliation{a}{Dept. of Nuclear Engineering and Radiological Sciences\\ University of Michigan, Ann Arbor, MI 48109}
\addAffiliation{b}{Physics Division, Los Alamos National Laboratory\\ Los Alamos, NM 87545}
\addAffiliation{c}{Dept. of Computer Science and Engineering \\ University of Michigan, Ann Arbor, MI 48109}

\addKeyword{Neutron Resonances}
\addKeyword{Machine Learning}
\addKeyword{Convolutional Neural Networks}
\addKeyword{SAMMY}

\titlePage

\begin{abstract}
This work investigates the feasibility of augmenting traditional R-Matrix codes with a robust machine learning framework for automatically detecting neutron resonances in transmission spectra. Neutron transmission data are often complex and noisy, making them difficult to analyze using traditional peak-identification methods. The state-of-the-art R-Matrix codes currently used by physicists to fit these data often depend on prior evaluations and require substantial manual effort. This preliminary study demonstrates a method for accelerating the post-experimental processing of neutron transmission data and reducing bias associated with dependence on prior evaluations. We employ a fully convolutional neural network to classify individual points as belonging to resonance or non-resonance regions in seven transmission spectra---two evaluated and five experimental. Although the model achieves classification accuracies in the range of 93\%, further analysis shows that this metric overstates its ability to generalize. Building on our original analysis in \cite{PHYSOR}, we find that, despite the inclusion of additional training data, the method does not generalize reliably to previously unseen isotopes. To address these limitations, future work should evaluate whether a larger and more diverse training dataset can produce a generalizable model and should incorporate known physical characteristics of neutron resonances to improve model performance.

\end{abstract}

\section{Introduction}\label{sec:intro}
Nuclear data underpin experiments, simulations, and reactor design across the nuclear enterprise. Inaccuracies in these data inflate uncertainties throughout the workflow. Advancing the field therefore requires improved measurement and, especially, robust statistical fitting. One challenge in nuclear data lies in fitting complex neutron resonances in energies where resonances can be experimentally resolved, the Resolved Resonance Region (RRR). These resonances are experimentally acquired via time-of-flight (ToF) transmission measurements, during which we can relate the time a neutron takes to travel a certain distance with its energy. Details on the experimental acquisition can be found in Ref. \cite{STAMATOPOULOS2022166166}. In ToF transmission, these resonances appear as irregular, narrow dips in the spectrum. Accurately resolving and fitting these resonances, the focus of this study, is essential for obtaining the energy-dependent cross sections relied upon by reactor designers, safety analysts, and code developers.

As it stands, analysts processing experimental data in the RRR typically rely on R-matrix evaluation codes---such as SAMMY \cite{nancym.larsonUpdatedUsersGuide2008} or REFIT \cite{moxon_1979}. While these codes are invaluable to the existing procedure of fitting transmission data, they carry some drawbacks for practicing physicists. First, these codes require prior distributions, necessarily skewing new calculations towards past evaluations. Additionally, these codes are prone to finding non-physical solutions to resonance parameter fitting when portions of the total resonance width are negligibly small. These errors require manual intervention and subjective correction, two obviously suboptimal steps for a procedure that demands precision. 

An ideal solution to this problem involves automatic detection of these high-frequency resonances that does not depend on a prior distribution and instead truly characterizes the experimental results at hand. Here it is important to note the critical impact R-matrix evaluation codes have on our understanding of nuclear data to date. Despite their intricate mathematics, both SAMMY and REFIT are white-box systems that should be augmented cautiously with black-box models like the ones described in this paper. In the interest of maintaining the scientific rigor of these well-documented systems, we do not propose a replacement of these methods with machine learning, but instead suggest using them in conjunction with efficient, ML-based peak-finding algorithms.

In this work, we apply a one-dimensional, fully convolutional neural network (1D FCN) to transmission spectra from experiments where neutrons in the meV-keV range impinge on a sample. As a brief background, during the impingement process, several reactions can occur: neutron capture (n,$\gamma$), elastic scattering (n,n), and fission (n,f). At characteristic energies, the cross section varies appreciably and we observe peaks, known as resonances. We describe resonances by a mean energy and a width, following a Breit-Wigner shape. For each type of reaction we have the corresponding widths---these are known as resonance parameters and are used to quantify cross sections for the aforementioned reactions. 

Following the automatic identification of the resonances, analysts could use a variety of techniques to obtain resonance parameters ($\Gamma_\gamma$ and $\Gamma_n$), whether that be through additional machine learning (ML) methods or through an augmented workflow in SAMMY or REFIT. This work presents the first phase of such an endeavor, automating the prediction of resonances within a transmission signal using a one-dimensional, fully convolutional network (FCN) that could subsequently be coupled with SAMMY for easier manual correction of misclassifications.

\section{Background}\label{sec:background}

\subsection{AI in Nuclear Data}

A natural solution to ``difficult pattern detection,'' as we find in neutron resonance identification, is machine learning. As such, several authors have attempted resonance/cross section prediction using ML in previous work. These works include a variety of attempts using ML to fit transmission spectra that either lie outside the resolved resonance region, focus on ``easier'' reaction types, or supplement R-matrix codes like SAMMY or REFIT, without actually identifying resonances themselves. Such works include fitting cross sections of photo-nuclear reactions with Bayesian neural networks \cite{sunEnhancingReliabilityPhotonuclear2025}, predicting (n, 2n) cross sections with regression trees \cite{teelock-gayaPredictionExoticN2n}, evaluating Pu-239 fission cross sections with ML to handle outlying or discrepant data \cite{whewellEvaluating239PunfCross2020}, predicting (n, 3n) cross sections using various ML models \cite{aliuncuPredictingN3nNuclear2025}, predicting ($\gamma$, n) cross sections using physics informed neural networks and gradient boosted decision trees \cite{besnard-vauterinExperimentalDatadrivenModeling2025},  predicting cross sections directly for a well-known isotope, U-235, using single phase shift neural networks \cite{huNovelDeepLearningbased2024}, and even wrapping SAMMY itself \cite{pleiades}. We also see analyses which use machine learning to classify quantum spin number, but this method depends on resonance energies and widths and does not use raw transmission data \cite{nobreNovelMachinelearningMethod2023}. Another effort to use ML to classify quantum spin number used decision trees and simulated data, and found a strong tendency to overfit \cite{brownMachineLearningApplied2020}. While all of these works use machine learning to predict cross sections, they are not generalizable methods for the resonance region of (n,$\gamma$), (n,n), or (n, f) reactions. Other attempts to capture nuclear cross-section uncertainties with deep neural networks were made by \cite{radaideh2021modeling}, reflecting the central role of nuclear data uncertainties across various reactor design applications, including reactor depletion \cite{radaideh2019advanced}, kinetics \cite{radaideh2019new}, and criticality safety \cite{price2019advanced,radaideh2018criticality}. Overall, our review indicates that the application of ML in the nuclear data field is still at an early stage of development.

The closest prior effort to our goal is Walton et al.'s automated resonance identification, which combines non-convex optimization with inferential statistics to infer resonance models directly from experimental ToF data with minimal manual intervention \cite{waltonAutomatedResonanceIdentification2024}. While this method is a major step towards reducing subjectivity in resonance fits, it still requires an analyst to employ SAMMY for their analysis. This project seeks to expand the work of Walton et al. to a generalizable method for resonance detection with future work aimed at resonance parameter characterization (e.g., $\Gamma_g, \Gamma_{\gamma}$). Xing et al. use a phase-shift deep neural network to predict resonance cross sections of the U-235 nucleus, but limit their analysis to a single isotope \cite{xingPhaseShiftDeep2024}. The resonance detection method we employ in this paper takes inspiration from peak detection via ML in other domains, with substantial adaptations to address the complex nature of nuclear transmission spectra. Some examples of peak detection via ML include convolutional neural network (CNN)-transformer models for arrhythmia detection \cite{kimNovelHybridCNNtransformer2025}, CNN classification and CNN integration for peak detection in raw liquid chromatography-mass spectroscopy \cite{melnikovDeepLearningPrecise2020}, and peak detection for chromatin immunoprecipitation followed by sequencing (ChIP-seq), a genomics identification problem that, much like neutron resonance structures, typically requires human oversight \cite{ohCNNPeaksChIPSeqPeak2020}.

\subsection{AI in Nuclear Engineering}

Artificial intelligence and machine learning have been applied across nearly every facet of nuclear power engineering, from reactor physics and safety analysis to plant operations and asset management. A substantial body of work has targeted surrogate modeling to ease the computational burden of high-fidelity simulations, beginning with the use of deep Gaussian processes as surrogates for advanced computer simulations \cite{radaideh2020surrogate}, building on earlier work combining simulation data with deep learning and uncertainty quantification \cite{radaideh2019combining,radaideh2020analyzing}. This line of work has since extended to seismic safety analysis of graphite reactor cores \cite{jones2022surrogate}, robotic radiation mapping via Gaussian process regression \cite{west2021use}, and high-dimensional BWR core neutronics modeling with deep neural networks \cite{saleem2020application}, with an automated machine learning platform developed to streamline such model deployment across nuclear applications \cite{myers2025pymaise}. Related efforts in predictive monitoring have used neural forecasting for loss-of-coolant accidents \cite{radaideh2020neural}, recurrent and convolutional LSTM autoencoders for anomaly detection in power electronics \cite{radaideh2022time}, and deep reinforcement learning to address class-imbalanced fault diagnosis \cite{zhong2023deep}. Reinforcement learning has further become central to autonomous reactor control, applied respectively to transient and load-following control \cite{tunkle2025nuclear}, multistep criticality search and power shaping \cite{radaideh2025multistep}, and physics-informed nuclear assembly design \cite{radaideh2021physics}, with its expanding role across nuclear energy systems synthesized in one review \cite{liu2024applications} and a second, complementary review \cite{gong2024possibilities}. More targeted studies have applied ML to control-sequence optimization for advanced reactors \cite{nguyen2024reinforcement}, operator support for monitoring safety functions \cite{park2020providing}, and adaptive maintenance-policy optimization under partial observability \cite{zhao2022reinforcement}.

AI-assisted optimization has also advanced core and fuel design, progressing from swarm and evolutionary algorithms for microreactor reactivity control \cite{price2022multiobjective}, to neuroevolution for large-scale BWR bundle design \cite{radaideh2021large}, to a convolutional neural network coupled with a genetic algorithm for PWR fuel-reloading pattern optimization \cite{wan2022optimization}. As such models grow more complex, interpretability has become a priority, addressed through symbolic regression with Kolmogorov-Arnold Networks aimed at opening the AI black box \cite{panczyk2025opening}. Digital twin technology, meanwhile, has become a major focus of nuclear AI research: a foundational framework first established the role of uncertainty in nuclear digital twins \cite{kochunas2021digital}, followed by an online autonomous calibration method for correcting low-fidelity digital twin simulations \cite{song2022online}, and more recently a variational digital twin framework introducing probabilistic reactor-state representation \cite{burnett2026variational}. At the operator-network level, deep neural operators first enabled real-time digital twin inference \cite{kobayashi2024deep}. At the system level, whole-system digital twins combining graph neural networks with System Analysis Module simulations have enabled full-state inference from sparse sensor data \cite{liu2025development}.

\section{Method}
\subsection{Data}
\label{sec:data}
In this analysis, we consider seven unique nuclear transmission datasets. These include a caesium-133 ENDF/B-VIII.0 evaluation \cite{brown_endfb-viii0_2018}, a samarium-149 ENDF/B-VIII.0 evaluation \cite{brown_endfb-viii0_2018} and experimentally acquired samarium-149, samarium-147, copper-63, iridium-191, and iridium-193 transmission spectra from the Device for Indirect Capture Experiments on Radionuclides (DICER), which is located at the Los Alamos Neutron Science C{E}nter (LANSCE) \cite{STAMATOPOULOS2022166166, DICER_neutronNews, DICER_NIC, DICER_NuclearSecurity, LAPPD, 88Zr_PRC, 88Zr_PRL}. 

Each dataset contains two input features: transmission, which ranges from zero to one, and energy. For this analysis, we chose to omit energy from our modeling scheme to allow our models to become ``independent'' of the target material(s) used in their training data without major deterioration in performance. Since the goal of this work is to use a well-trained model to predict resonance ranges on an unseen energy spectrum, we create three unique models, each trained on four of the transmission datasets described above, and each tested on the two withheld spectra (we exclude the samarium-149 evaluation file from these cases). This strategy yields three unique cases to evaluate our method on. However, since these spectra differ dramatically, we also experimented with training and testing on the same spectrum, as shown in our original conference proceedings \cite{PHYSOR}. We encourage future work to replace both our case-based approach and single-spectrum approach with one robust model trained on many more datasets. Table \ref{tab:case_outline} shows how we use the datasets mentioned above in each case. 

\begin{table}[!ht]
    \centering
    \caption{Case descriptions for the three generalized models and three single spectrum models from \cite{PHYSOR}.}
    \begin{tabular}{lrr}
        \toprule
         \textbf{Case}  & \textbf{Training Isotopes} & \textbf{Testing Isotopes}  \\
         \midrule
            1 & Cs-133*, Ir-193, Ir-191, Sm-149 & Cu-63, Sm-147 \\
            2 & Cu-63, Sm-147, Ir-191, Sm-149 & Cs-133*, Ir-193 \\
            3 & Cu-63, Sm-147, Cs-133*, Ir-193 & Ir-191, Sm-149 \\
            A & Cs-133* & Cs-133* \\
            B & Sm-149* & Sm-149* \\
            C & Sm-149 & Sm-149 \\
         \bottomrule
    \multicolumn{3}{l}{\small *Represents an evaluation dataset.} \\
    \end{tabular}
    \label{tab:case_outline}
\end{table}

As with most applications of machine learning, we anticipate more data to improve the results of this paper and therefore present our findings as a preliminary study on using a one-dimensional FCN to predict resonance regions of neutron transmission spectra. 

\label{sec:method}
\subsection{Pre-Processing}
The method we employ in this work requires a multi-step pre-processing scheme. 
First, since convolutional neural networks (CNNs) require evenly spaced samples, we load the experimental transmission dataset of interest and fit the noisy spectrum precisely using B-splines via the Scipy interpolation function \verb|make_interp_spline()|. We verify the precision of this fit by evaluating the residuals between the original and fit data points, which all are on the order of 1e-16. Then, we evaluate the generated spline curve as a function at 256,000 evenly-spaced energy values to yield 256,000 transmission samples. While this number of samples is arbitrary, this step ensures each dataset has an equal number of samples after pre-processing, independent of the energy ranges of the transmission spectra. Because the energy ranges of the transmission spectra vary, this implies that some datasets will have more closely spaced points than others, which may affect the performance of the model. By choosing a relatively large number of data points, we aimed for a ``continuous'' spectrum to mitigate performance gaps caused by substantial jumps between data points. 

\begin{figure}[ht!]
    \centering
    \includegraphics[width=\textwidth]{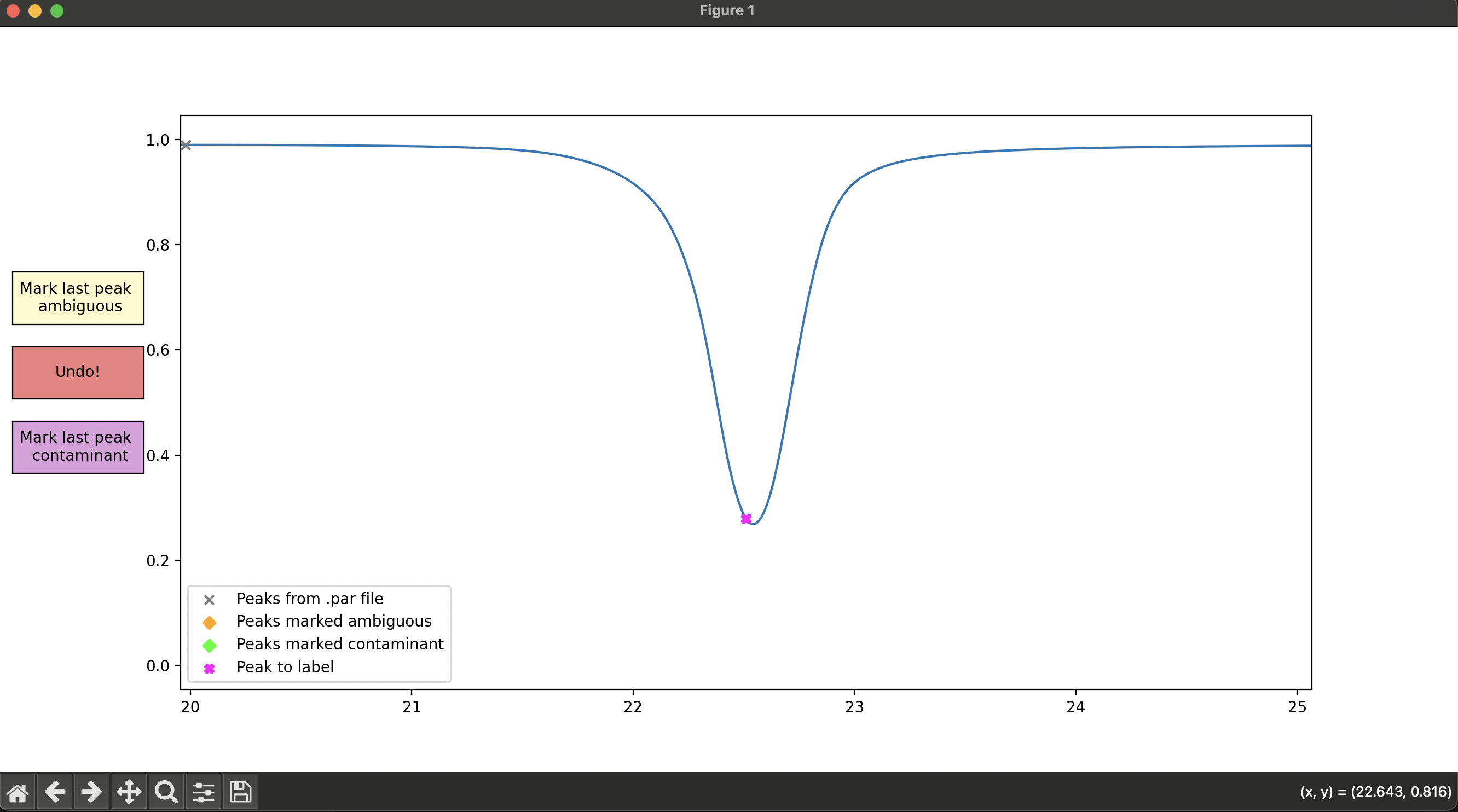}
    \caption{Example shot of interactive labeling tool on an evaluation spectrum.}
    \label{fig:demo_selection}
\end{figure}

After sampling our spline curves, we label the peak regions of the data. To do this, we implemented an interactive selection tool that allows a user to select the start and stop of each peak in the dataset. This tool uses an additional input file, a parameter file from SAMMY, to mark the expected peak locations on the uniformly sampled transmission spectra, as shown in Figure \ref{fig:demo_selection}. 

The parameter file allows for automatic zooming and advancing, but could be substituted or removed pending further development of the tool. As a user selects a region, the labeling tool writes a corresponding label for each point in the uniformly sampled distribution. Data points that reside within a user-defined ``peak'' correspond to a ``1'' in the label file. Likewise, data points that lie outside regions of interest are marked ``0'' in the label file. Optionally, and likely depending on the complexity of the spectrum, the user may mark contaminant regions as ``2'' and ambiguous regions of ``3.'' We provide these additional labeling options to allow for both multi-class modeling efforts in future work and to track regions of uncertainty. For the purposes of this analysis, we set our models to perceive all values of ``2'' or ``3'' as ``0'' or ``non-peak'' regions.

\begin{figure}[ht!]
    \centering
    \includegraphics[trim=10pt 10pt 10pt 10pt, clip, width=\textwidth]{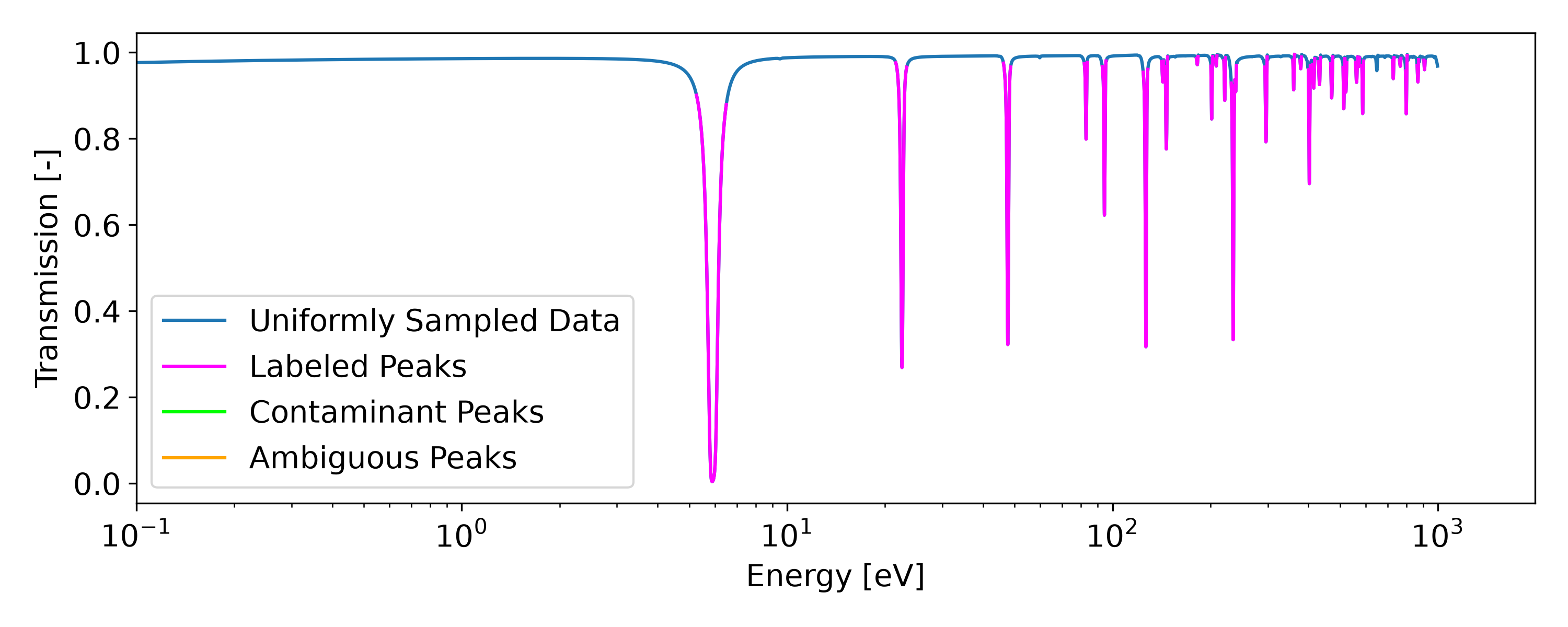}
    \caption{Caesium-133 evaluation labeled transmission spectrum.}
    \label{fig:cesium-labels}
\end{figure}

\begin{figure}[ht!]
    \centering
    \includegraphics[trim=10pt 10pt 10pt 10pt, clip, width=\textwidth]{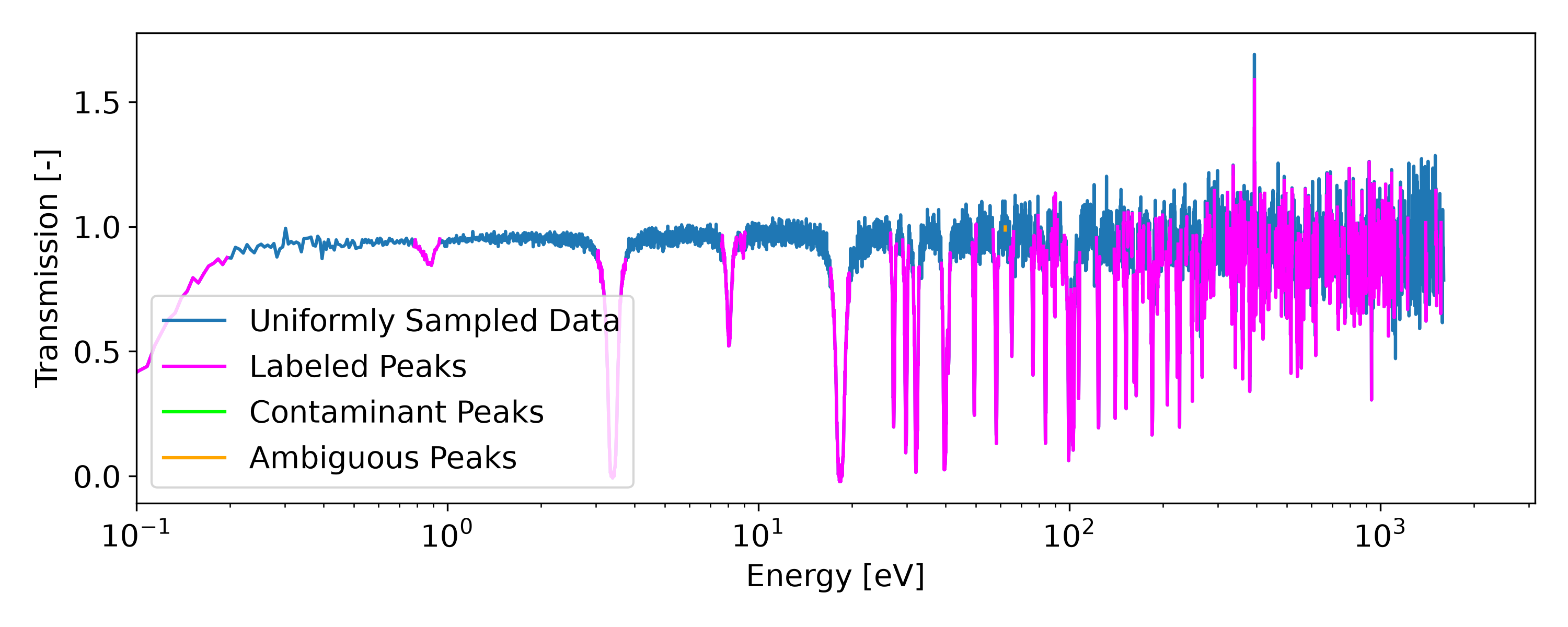}
    \caption{Experimental samarium-149 labeled transmission spectrum.}
    \label{fig:samarium-exp-labels}
\end{figure}

Figures \ref{fig:cesium-labels} and \ref{fig:samarium-exp-labels} show examples of labeled transmission spectra for the caesium-133 evaluation and samarium-147 experimental datasets, respectively. To simplify the labeling process for this workflow, we did not track ambiguous or contaminant peaks. It is also important to note that the actual accuracy of the peak labels (how we labeled them vs the physically true peak start-stop indices) in Figures \ref{fig:cesium-labels} and \ref{fig:samarium-exp-labels} is not critical to the findings of this paper. The objective of this work is to explore how well a 1D FCN can replicate the job of a human and R-matrix code in identifying the resonances within a transmission spectrum, we expect the model to perform even better with more, and more precisely labeled data. We also note that the lower energy ranges across this analysis have a higher density of true peaks than the higher energy ranges. This is because the peaks are more distinguishable at lower energies where they have a higher amplitude and are therefore labeled more frequently, but leads to consequences in model performance.

After labeling each dataset, our datasets contain a single input feature: transmission, and a single output feature: a label, on a per-point basis. We then need to split up the datasets into a number of sequences. That exact number is another tunable parameter that we arbitrarily set to 1,000 points/samples per sequence. Before dividing the full dataset into sets of 1,000, we first shuffle all the pairs of transmission values and corresponding labels, such that the sequences are discontinuous in energy. This helps provide full-spectrum context during training and extends the flexibility of our limited data. Here, we allow an additional tunable parameter of ``overlap,'' which would include some number of repeat samples in each sequence. For simplicity, we excluded overlap in this analysis, but note that future work may find this feature beneficial. Finally, since our transmission values range from 0-1, we do not scale our datasets. If a model were to be trained on and applied to single isotope, i.e., a model trained on multiple experimental transmission spectra of the same isotope and then used to make predictions based on additional experimental data of that same isotope, then energy should be included as a second feature and should be scaled accordingly. 

For the three generalized models, we append the pre-processed training data arrays together to create one large input dataset for each generalized model.  Excluding energy makes this possible. We do not append the test datasets together, however, as we evaluate them individually.

\subsection{Model Architecture and Training}
To make our peak predictions, we employ a fully convolutional network (FCN)---a CNN without a fully connected layer. By excluding the fully connected layer, our model can make point-wise classification predictions. The nature of our datasets requires this flexibility because the peaks are not evenly spaced or identical in structure, so we cannot ask the model to predict if a given sequence contains a peak, rather, we ask what class each ``pixel'' (transmission value) belongs to. This flexibility, however, comes at the cost of introducing room for error. Since the goal of this detection algorithm is not to find the start and stop of the resonance, but merely its peak location, by making predictions on a pixel-wise basis, we ask for more precision than is required by our application. We will discuss the implications of this phenomenon later in the analysis.

\begin{figure}[ht!]
    \centering
    \includegraphics[width=0.75\textwidth]{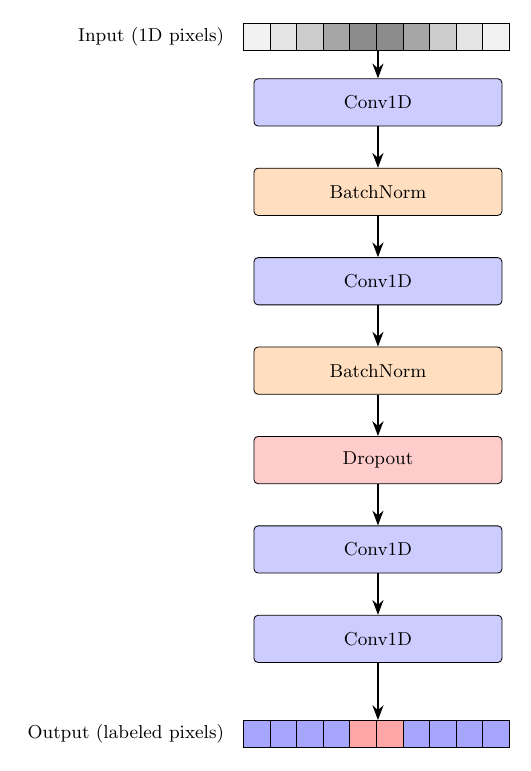}
    \caption{Modeling architecture schematic.}
    \label{fig:schematic}
  \end{figure}

Notwithstanding the lack of a fully connected layer, our model architecture is one of a fairly typical CNN. It includes two alternating sets of 1D convolutional layers and batch normalization layers, then a dropout layer, followed by two more 1D convolutional layers. Figure \ref{fig:schematic} shows this structure, including a snippet of what the input and output data structures look like. 

\begin{figure}[ht!]
    \centering
    \includegraphics[width=0.75\textwidth]{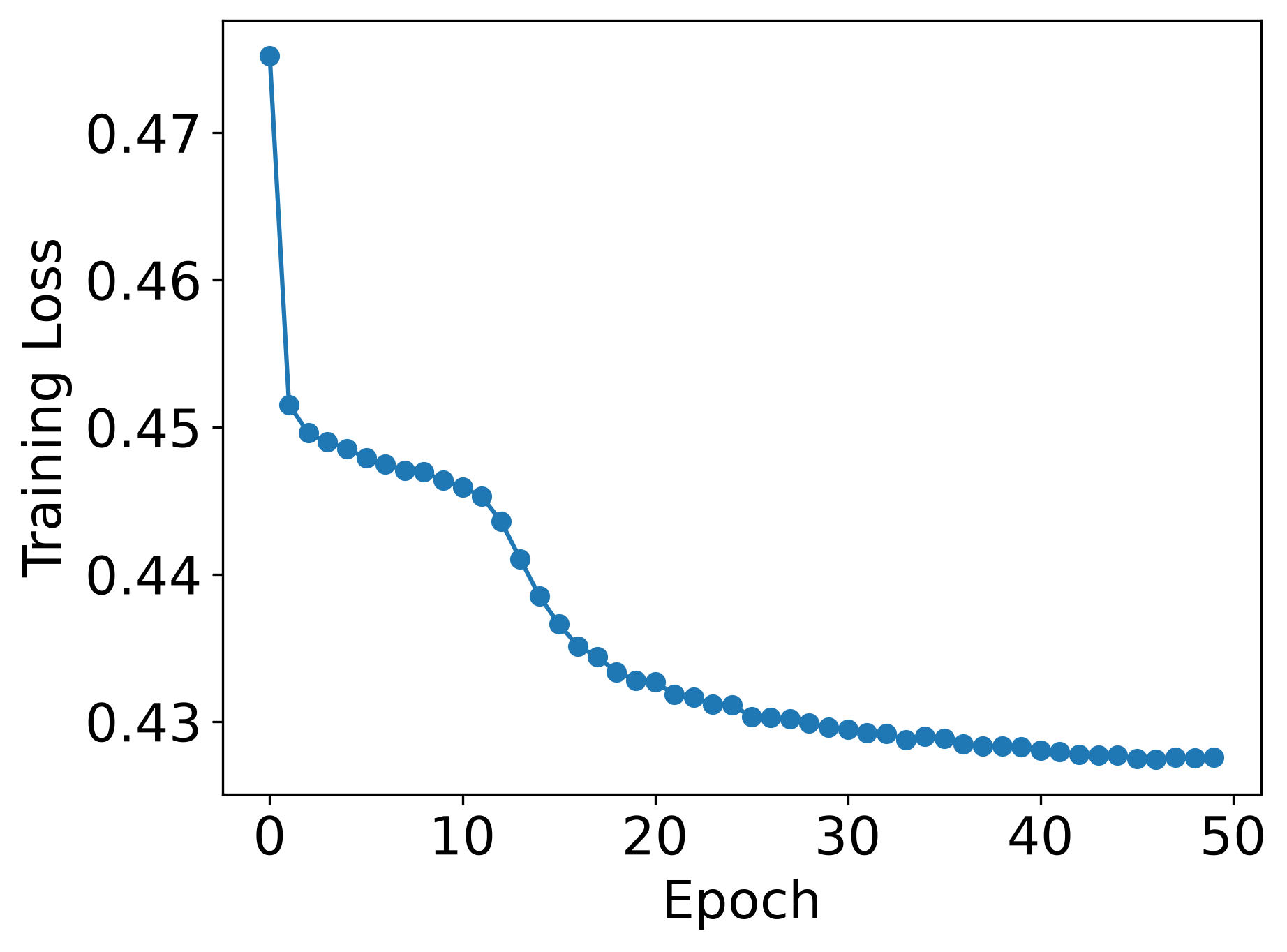}
    \caption{Iridium-191 loss curve.}
    \label{fig:loss_curve}
\end{figure}

We train each model for a default of 50 epochs for all cases of models using Adam as our optimizer, binary cross entropy with logits loss as our loss function, and cosine annealing as our learning rate scheduler. We apply a positive weight of 1.33 (the ratio of negative to positive samples in one of the least skewed datasets, samarium-149 experimental from \cite{PHYSOR}) to our loss function to accommodate class imbalance in a uniform way across datasets. We found 50 epochs sufficient to reach a plateau on the loss curves for each model. Figure \ref{fig:loss_curve} shows an example of the iridium-191 loss curve.

We describe the F1 score, along with other metrics used in our analysis (precision and recall), in Equations \ref{eq:precision}--\ref{eq:f1}.
\begin{equation}
    Precision = \frac{TP}{TP + FP}
    \label{eq:precision}
\end{equation}
\begin{equation}
    Recall = \frac{TP}{TP + FN}
    \label{eq:recall}
\end{equation}
\begin{equation}
    F1 = 2 \times \frac{Precision \times Recall}{Precision + Recall}
    \label{eq:f1}
\end{equation}
where $TP$ is a true positive, $FP$ is a false positive, and $FN$ is a false negative. In this analysis we employ a 0.5 threshold for classification. In other words, pixels with a probability greater than 0.5 of being a ``peak'' will count as peaks. 

We use the metrics above because they are standard for machine learning classification tasks, however, they are far from sufficient for evaluating our models. As shown in Figure \ref{fig:schematic}, the model makes pixel-wise predictions due to the lack of the fully connected layer that is standard in most CNNs. Because the prediction for each of the individual pixels get factored into the metrics, this means that the metrics are equally sensitive to correct predictions over the full range of the peaks. However, for this application, we need high accuracy at the center of the peak and have flexibility at the base (start and stop regions). This means that the metrics we present in this study may deflate their quality for the desired application. 

\subsection{Hyperparameter Tuning}
To optimize its performance, we hyperparameter tune our FCN for each of the cases. For each generalized case's hyperparameter tuning, we train and test using two k-folds and calculate the F1 score on both folds, taking the average as our criterion for optimization. For the single-spectrum analyses completed in \cite{PHYSOR}, we used five k-folds to account for the smaller dataset size. During hyperparameter tuning for the generalizable cases (1-3), we use a split of the ``training'' datasets for each case, randomly shuffled across isotopes, i.e, we do not introduce the actual test isotopes during tuning. We optimize using a Tree-Structured Parzen Estimator algorithm (TPE) as defined by the \verb|Optuna| package. The CNN parameters tuned include the number of filters in each layer, kernel size, dropout rate, batch size, and the initial learning rate, where a representative range is selected for each parameter.

\section{Results and Discussion}
\label{sec:results}

Table \ref{tab:hyperparameter_results} shows the best hyperparameter tuning results for each of the generalized cases (Cases 1-3). Table \ref{tab:hyperparameter_results_physor}, shown here as in \cite{PHYSOR}, presents the best hyperparameters found for the single-spectrum cases (Cases A-C).

\begin{table}[!ht]
    \centering
    \small
    \caption{Best hyperparameters found after 50 evaluations on each dataset using two epochs and two k-folds per trial.}
    \begin{tabular}{lrrr}
        \toprule
         \textbf{Hyperparameter}  & \textbf{Case 1} & \textbf{Case 2} & \textbf{Case 3}  \\
         \midrule
         Filters & 128 & 128 & 16\\
         Kernel Sizes & [7, 5, 9, 3] & [5, 5, 5, 9] & [3, 7, 9, 3]\\
         Dropout Rate & 0.41 & 0.21 & 0.29 \\
         Batch Size & 16 & 16 & 16\\
         Initial Learning Rate & 0.00849 & 0.00856 & 0.00997\\
         \midrule
         Total Trainable Parameters & 396,547 & 331,267 & 8,547 \\
         \bottomrule
    \end{tabular}
    \label{tab:hyperparameter_results}
\end{table}

\begin{table}[!ht]
    \centering
    \small
    \caption{Best hyperparameters found in \cite{PHYSOR} using 100 evaluations on each dataset with two epochs per trial.}
    \begin{tabular}{lrrr}
        \toprule
         \textbf{Hyperparameter}  & \textbf{Case A} & \textbf{Case B} & \textbf{Case C}  \\ 
         \midrule
         Filters & 128 & 64& 32\\
         Kernel Sizes & [7, 9, 5, 7] & [3, 5, 5, 3] & [5, 5, 3, 9]\\
         Dropout Rate & 0.33& 0.21 & 0.31 \\
         Batch Size & 16 & 16 & 16\\
         Initial Learning Rate & 0.00517 & 0.00468 & 0.00264\\ 
         \midrule
         Total Trainable Parameters & 461,825 & 82,945 & 17,153 \\
         \bottomrule
    \end{tabular}
    \label{tab:hyperparameter_results_physor}
\end{table}

\begin{table}[h]
    \caption{Model metrics for each test isotope across all three generalized cases and all three single single-spectrum cases from \cite{PHYSOR}.}
    \centering
    \begin{tabular}{lrrrrrr}
        \toprule
    Case & Test Isotope & Average Loss & Accuracy & Precision & Recall & F1 Score \\
    \midrule
    \multirow{2}{*}{1} & Cu-63 & 0.333 & 0.908 & 0.469 & 0.368 & 0.412 \\
                       & Sm-147 & 0.670 & 0.753 & 0.823 & 0.182 & 0.298 \\
    \hline
    \multirow{2}{*}{2} & Cs-133* & 0.393 & 0.857 & 1.000 & 0.037 & 0.071 \\
                       & Ir-193 &0.331 & 0.908 & 0.455 & 0.389 & 0.419 \\
    \hline
    \multirow{2}{*}{3} & Ir-191 & 0.313 & 0.937 & 0.729 & 0.269 & 0.393 \\
                       & Sm-149 & 0.802 & 0.641 & 0.977 & 0.232 & 0.375 \\
    \midrule
    \multirow{1}{*}{A} & Cs-133* & 0.141 & 0.955 & 0.847 & 0.845 & 0.846 \\
    \hline
    \multirow{1}{*}{B} & Sm-149* & 0.348 & 0.875 & 0.846 & 0.831 & 0.839 \\
    \hline
    \multirow{1}{*}{C} & Sm-149 & 0.541 & 0.780 & 0.750 & 0.733 & 0.741 \\

    \bottomrule
    \multicolumn{4}{l}{\small *Represents an evaluation dataset.} \\
    \end{tabular}
    \label{tab:all_results}
\end{table}

Overall, Table \ref{tab:all_results} shows clearly superior performance across all metrics on the single-spectrum results. This outcome, though unexpected, is not surprising-- the single-spectrum cases saw a limited amount of training data, but it was very similar to the test data (as a result of being from the same spectrum). The generalized cases saw more than four times as much training data, but none from the isotopes they were asked to predict. We hypothesized the dramatic increase in training data to more than accommodate for the lack of \textit{similar} training data, but as shown by the results in Table \ref{tab:all_results}, this was not observed.

\begin{figure}[htbp]
    \centering
    \begin{subfigure}[b]{0.91\textwidth}
        \centering
        \includegraphics[width=\textwidth]{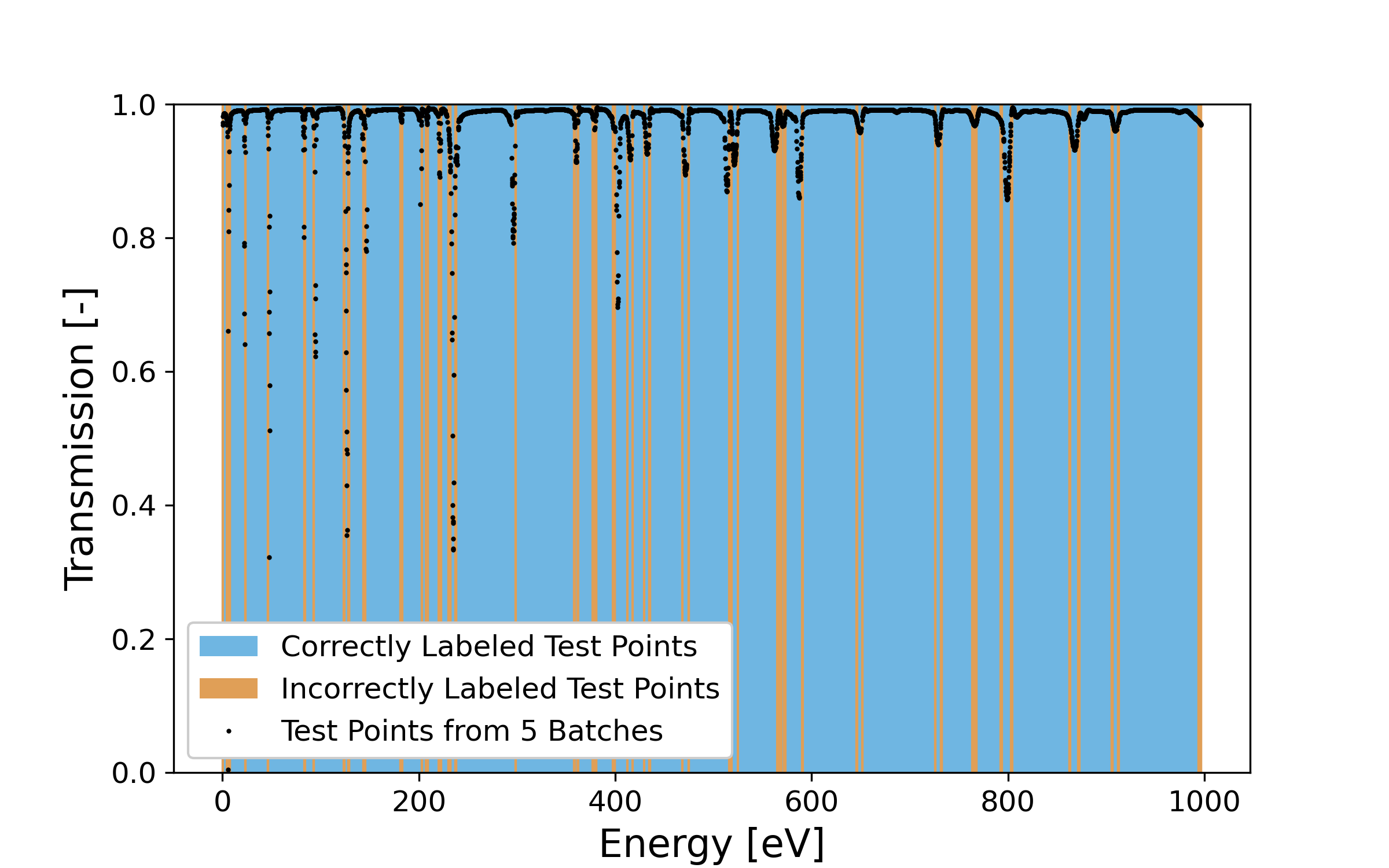}
        \caption{Single-spectrum (Case A) predictions for five batches of caesium-133.}
        \label{fig:flags_cs133singlespec}
    \end{subfigure}
    \vspace{1em}
    \begin{subfigure}[b]{0.91\textwidth}
        \centering
        \includegraphics[width=\textwidth]{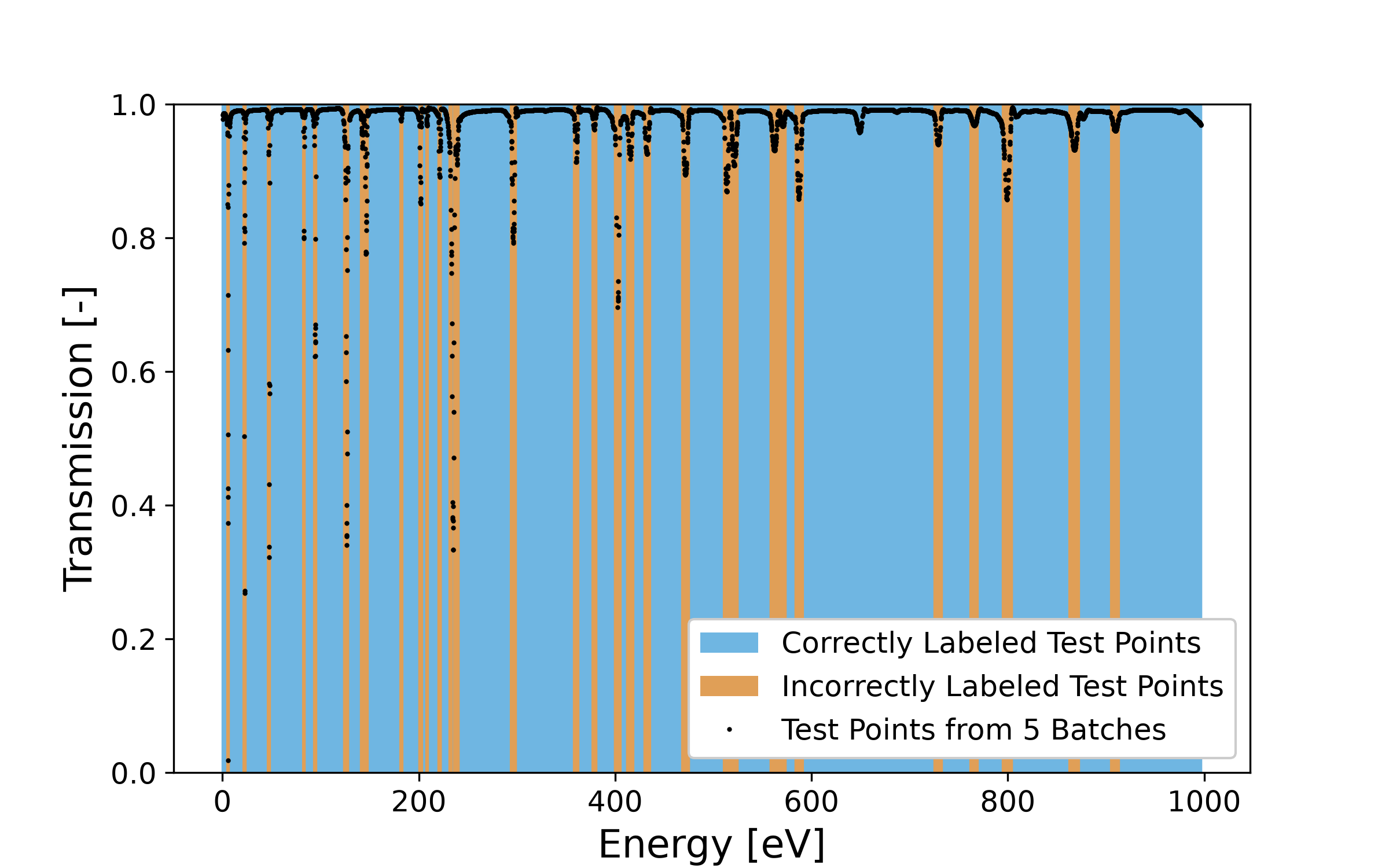}
        \caption{Generalized model (Case 2) predictions for five batches of caesium-133.}
        \label{fig:flags_cs133general}
    \end{subfigure}
    \caption{Comparison of caesium-133 predictions from single-spectrum vs generalized model.}
    \label{fig:cs133preds}
\end{figure}

Upon further inspection, the results of the single-spectrum cases are better than they appear, while the results of the generalized models are worse. For a straightforward comparison, we show the predictions of the simplest spectrum tested in this analysis-- the caesium-133 evaluation dataset. Figure \ref{fig:cs133preds} shows the predictions for both models on the caesium-133 dataset over five batches of data. 

Looking closely at Figure \ref{fig:cs133preds} highlights the discrepancy in the metrics shown in Table \ref{tab:all_results} between these two models. For clarity, the visualization we present in Figure \ref{fig:cs133preds} plots five batches of data points (five samples) corresponding with their original energy values. Then, we plot a vertical blue or orange band over each point, depending on whether the class of that point was predicted accurately by the model. Because the points and vertical bands are both larger than the true size of the points in the model, the overlap may cause a slight distortion of the actual results. Nonetheless, we find this style of visualization critical for understanding qualitative details about the model performance that metrics alone cannot provide.

In the single spectrum cases (both caesium-133 and samarium-149), the model frequently makes incorrect predictions on the edges of the peaks. This is shown by two orange bands on the edges of each peak (marking an incorrect prediction) with blue bands in the center region of the peak (marking a correct prediction) in Figure \ref{fig:flags_cs133singlespec}. This visualization therefore shows that most of the error in this model comes from the uncertainty around the start and stop of the peak. However, because the exact start and stop locations are not critical to the application (we are mostly interested in the center location of the peak), this error is relatively unproblematic. In fact, this visualization strengthens the performance evaluation of the single-spectrum case-- it is accurate precisely where needed.

\begin{figure}[htbp]
    \centering
    \begin{subfigure}[b]{0.9\textwidth}
        \centering
        \includegraphics[width=\textwidth]{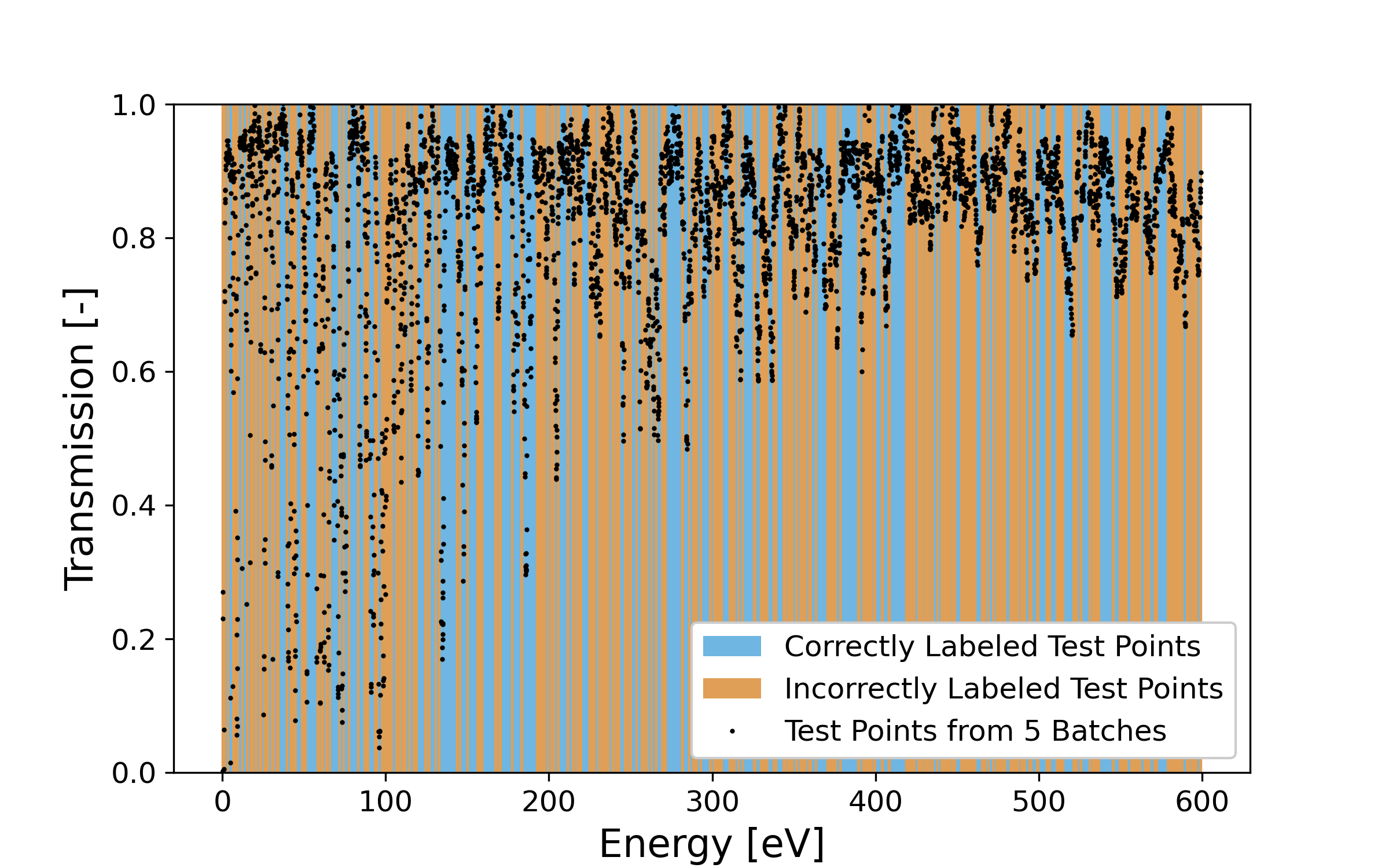}
        \caption{Single-spectrum (Case C) predictions for five batches of samarium-149.}
        \label{fig:flags_sm149singlespec}
    \end{subfigure}
    
    \vspace{1em}
    
    \begin{subfigure}[b]{0.9\textwidth}
        \centering
        \includegraphics[width=\textwidth]{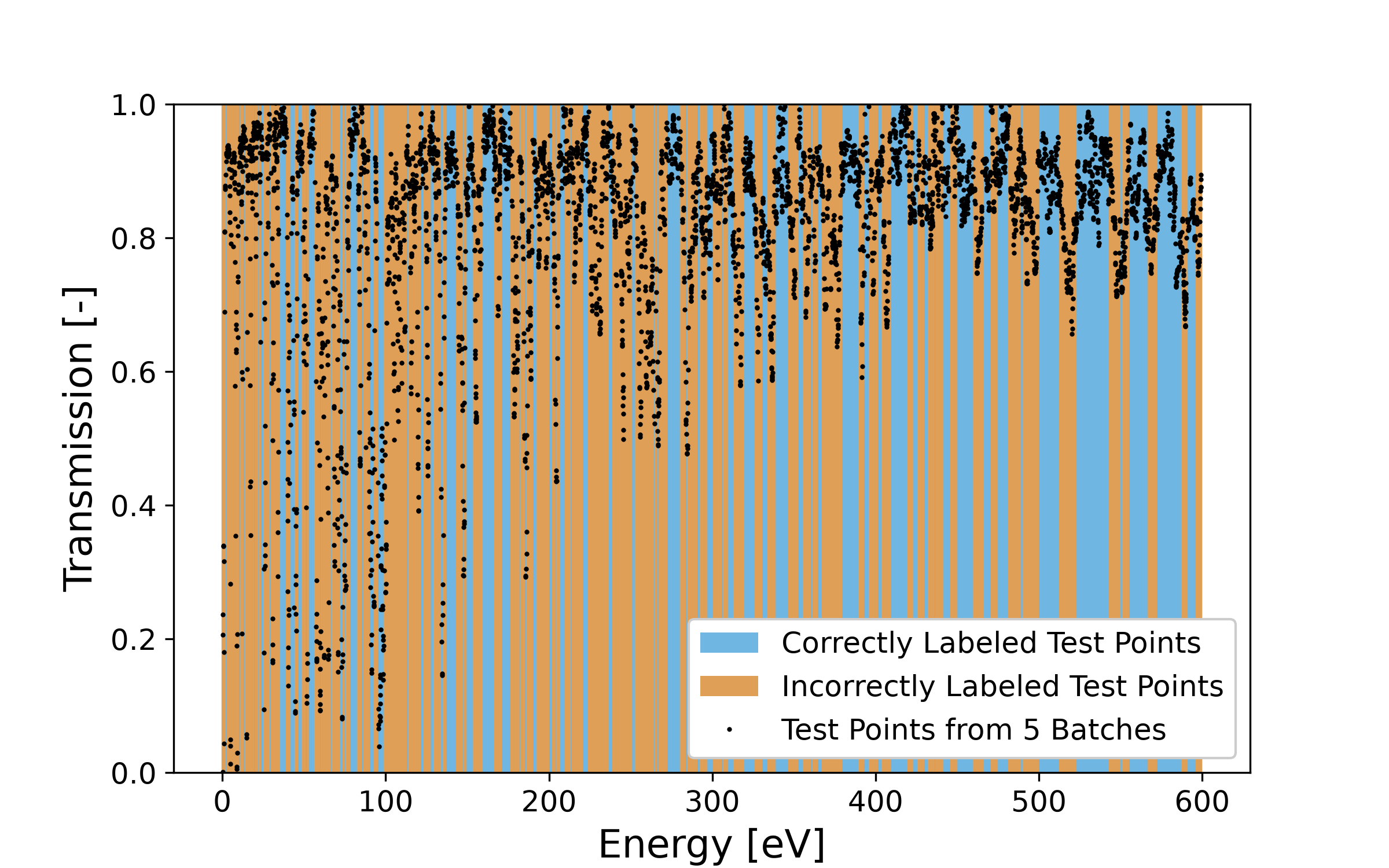}
        \caption{Generalized model (Case 3) predictions for five batches of samarium-149.}
        \label{fig:flags_sm149general}
    \end{subfigure}
    \caption{Comparison of samarium-149 predictions from single-spectrum vs generalized model.}
    \label{fig:sm149preds}
\end{figure}

In the generalized case, shown in Figure \ref{fig:flags_cs133general}, the model misses nearly every peak. Yet, the generalized case still has decent accuracy, as shown in Table \ref{tab:all_results}. The accuracy and precision metrics in Table \ref{tab:all_results} prove misleading-- the model circumvents the task of deciphering the peaks from non-peaks by overwhelmingly labeling regions as the majority class-- non-peak. The model suffers performance losses in terms of recall, which is later reflected in the F1 score, but maintains relatively high accuracy. This behavior could be attributed to a weak class imbalance weight during training, since we based this on the samarium-149 experimental dataset, which has many more peak regions than caesium-133 and other cases. However, a quick look at the samarium-149 metrics shown in Table \ref{tab:all_results} and the visualization shown in Figure \ref{fig:sm149preds} shows that, regardless, a more accurate weighting factor fails to overcome as complex of a spectrum as samarium-149. Again, the proximity of the precision (0.750) and recall (0.733) scores for the single spectrum samarium-149 case in Table \ref{tab:all_results} compared to the gap between these two values in the generalized case-- 0.977 vs 0.232 for precision and recall, respectively, shows that the latter model favors a single class label (non-peak) and suffers the consequence for its recall metrics. We attempted to address this in model training by optimizing over the F1 score, which balances recall and precision, but this proved fruitless for the general case.

We found caesium-133 and samarium-149 to be good demonstrative cases for this analysis-- comparing the ``easiest'' and ``hardest'' spectra, respectively. However, we present the visualizations for the remaining generalized cases in Appendix \ref{general_viz}. We largely observe worse performance at lower energies across these cases, however, this is because there are more labeled peaks due to their greater amplitudes at lower energies. Likewise, the models appear to perform better at higher energies, but this is because the models prefer non-peak regions, and these higher energies have a greater density of non-peaks.

\section{Conclusions and Future Work}
Overall, as a feasibility study, the original analysis we conducted \cite{PHYSOR} demonstrated high potential for successful future work. Even with some limited testing on unseen spectra in our original study, we observed good model performance \cite{PHYSOR}. Counterintuitively, the addition of more spectra to the training set in this analysis appeared to degrade the model's ability to accurately distinguish peaks. This could stem from adding too much variety in the broader training data set without sufficient repetition, the difficulty of this task being underscored by the noisy nature of the data. It could also mean that a physics-agnostic method is unsuitable for this problem. R-matrix codes like SAMMY account for decades of refined physics that exceeds the capacity of pure pattern recognition, at least with the method and datasets applied in this analysis. To this end, we propose that future work embed some physics-based modeling into a machine learning method, like the one applied here. It is also possible that resonance fitting will require exceptionally narrow (as opposed to general) AI. We see successful single isotope analyses like this in the literature \cite{xingPhaseShiftDeep2024} and in our initial efforts \cite{PHYSOR}. Future work should further assess whether more datasets enable 1D CNNs to accurately capture resonance patterns, or if the problem remains persistent. These datasets may include additional experimental data or supplementation with synthetic data, such as from diffusion models.

If future work successfully resolves the issues found in this analysis, an ideal next step would package the method to run a baseline model in an automated regime at DICER and other stations. Future work may find more advanced models, particularly those with some memory, to be less sensitive to the peak boundary issue found in \cite{PHYSOR}. Ultimately, the goal of this research should extend the benefits of machine learning to the efficacy of R-matrix codes. While this extension of our original analysis \cite{PHYSOR} disproved our hypothesis that additional data would immediately improve the ability of our models to generalize, we exposed complexities of the problem to help guide future research. We expect further efforts to find solutions to the shortcomings of our method and help improve the overall resonance fitting workflow.

\pagebreak
\section*{Acknowledgments}
The first author of this work is supported by the National Science Foundation's Graduate Research Fellowship Program (DGE-2241144). This work was supported by the U.S. Department of Energy through the Los Alamos National Laboratory. Los Alamos National Laboratory is operated by Triad National Security, LLC, for the National Nuclear Security Administration of U.S. Department of Energy (Contract No. 89233218CNA000001). This work was also supported in part by the Nuclear Criticality Safety Program, funded and managed by the National Nuclear Security Administration for the Department of Energy.

\pagebreak
\bibliographystyle{style/ans_js}
\bibliography{bibliography}

\FloatBarrier

\appendix

\section{Visualizations from Generalized Models}
\label{general_viz}

\begin{figure}[ht!]
    \centering
    \includegraphics[width=\textwidth]{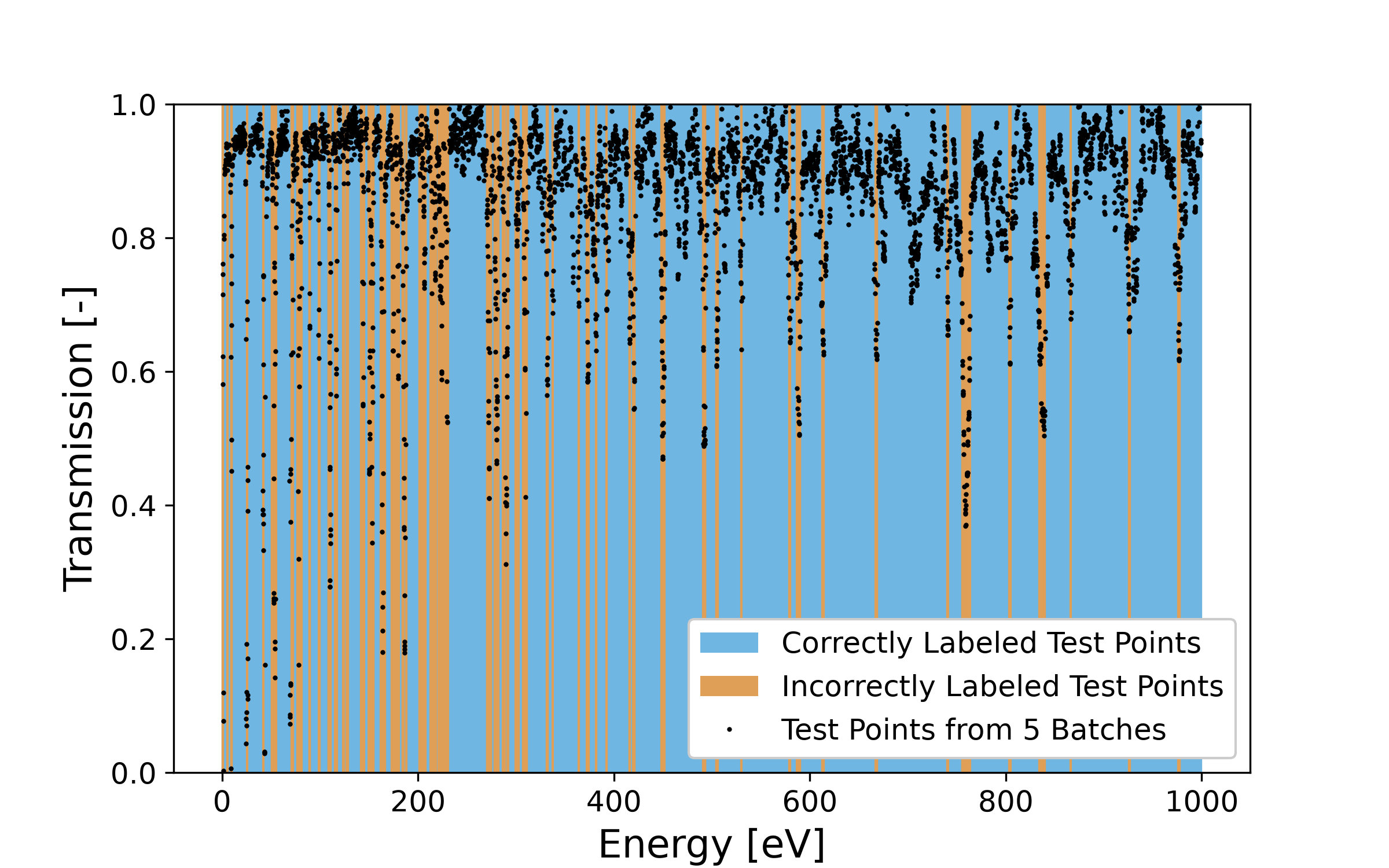}
    \caption{Prediction results for Cu-63 on the Case 1 general model.}
    \label{fig:cu63_preds}
\end{figure}

\begin{figure}[ht!]
    \centering
    \includegraphics[width=\textwidth]{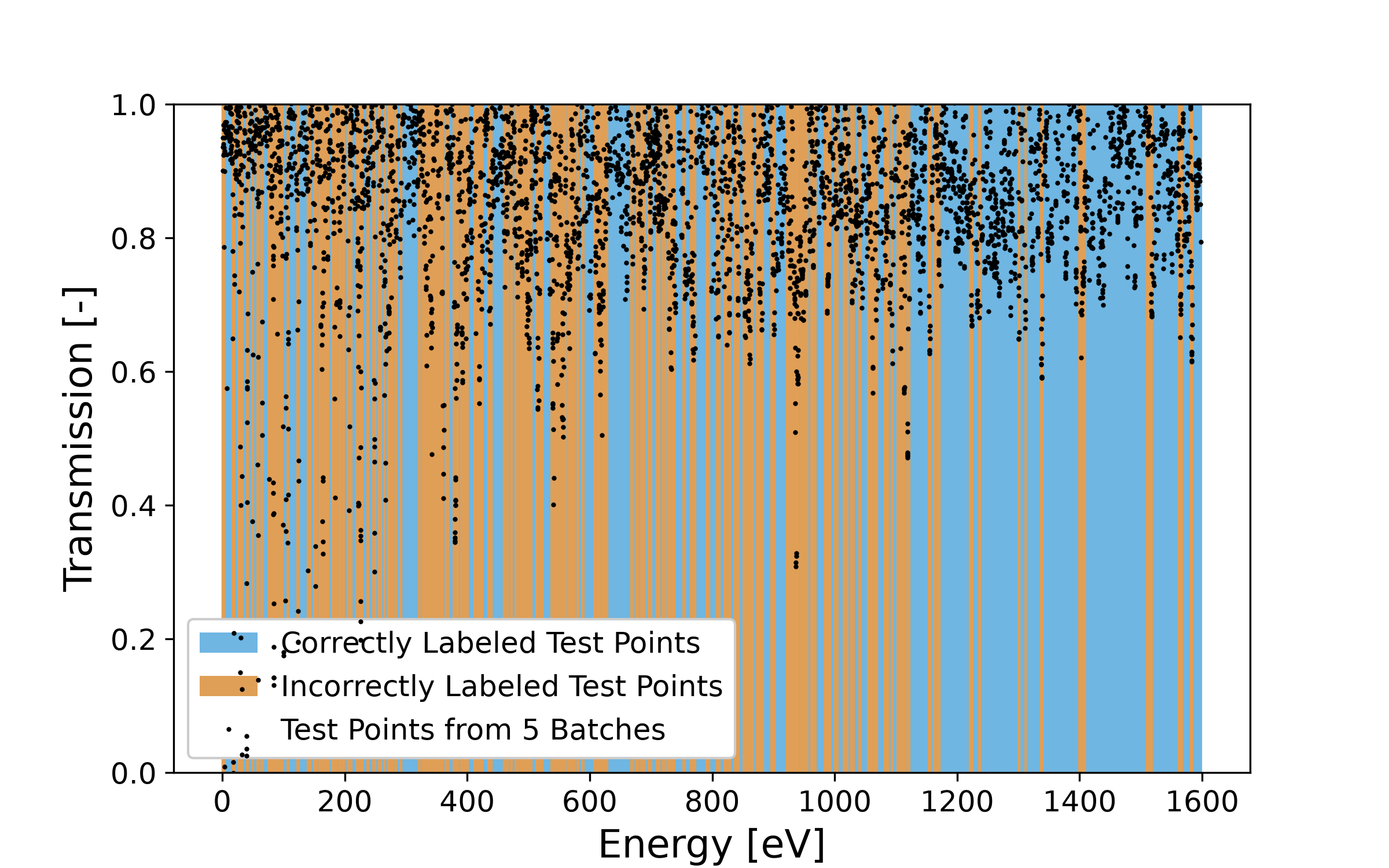}
    \caption{Prediction results for Sm-147 on the Case 1 general model.}
    \label{fig:sm147_preds}
\end{figure}

\begin{figure}[ht!]
    \centering
    \includegraphics[width=\textwidth]{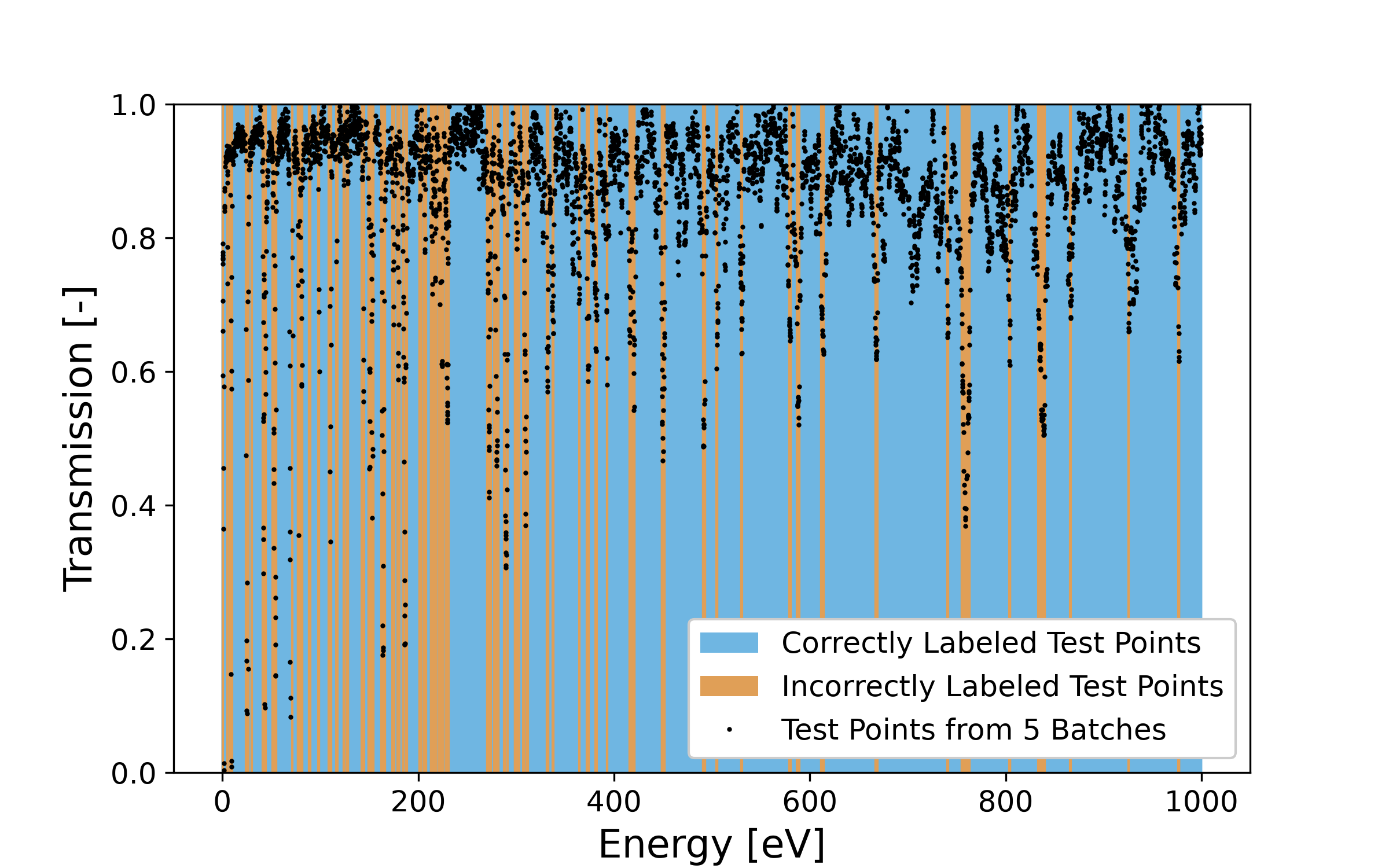}
    \caption{Prediction results for Ir-193 on the Case 2 general model.}
    \label{fig:ir193_preds}
\end{figure}

\begin{figure}[ht!]
    \centering
    \includegraphics[width=\textwidth]{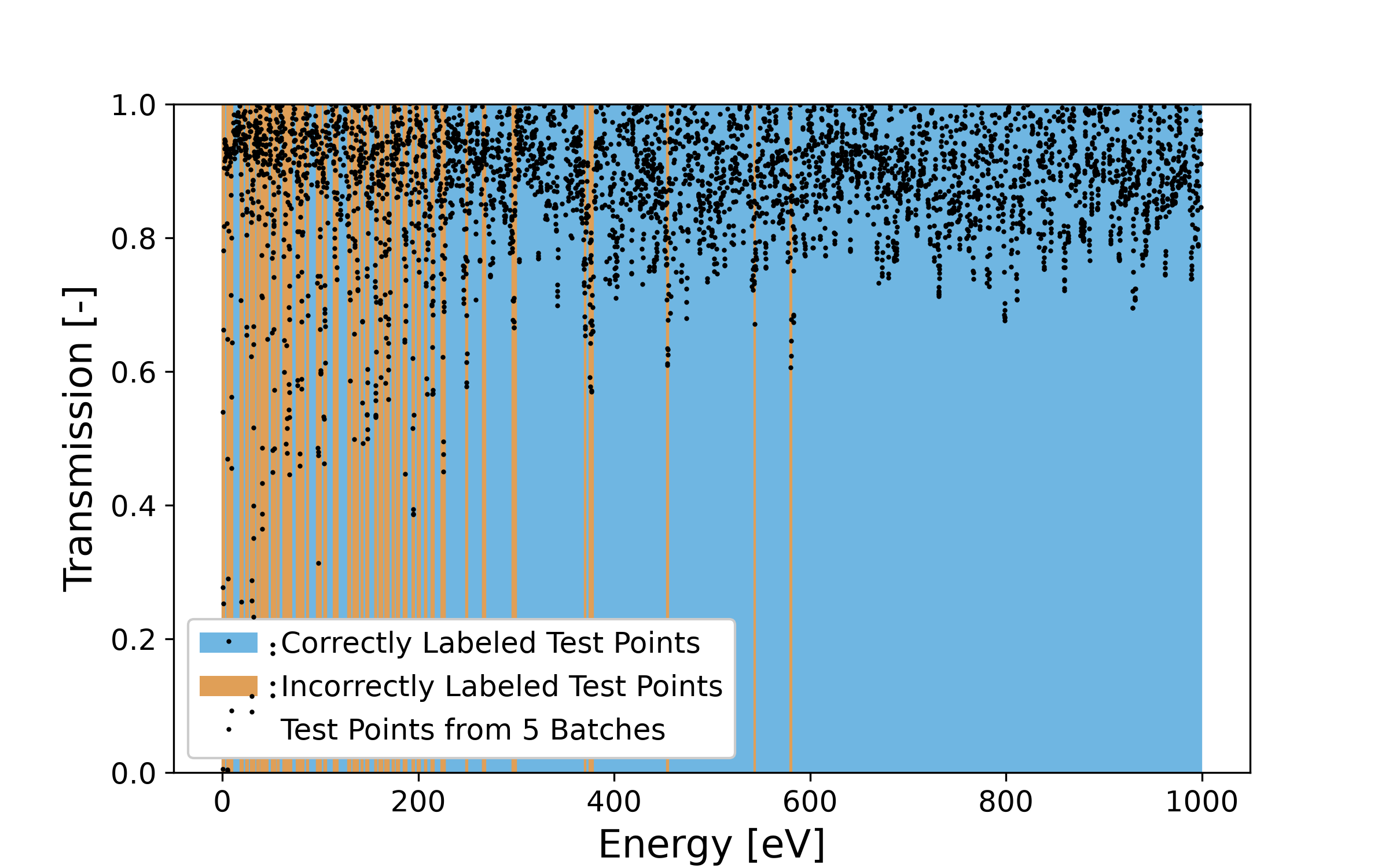}
    \caption{Prediction results for Ir-191 on the Case 3 general model.}
    \label{fig:ir191_preds}
\end{figure}

\end{document}